# Towards Trustable Language Models: Investigating Information Quality of Large Language Models

**Rick Rejeleene**
rrejeleene@ualr.edu
Department of Information Science
University of Arkansas
Little Rock 72204

**Xiaowei Xu**
xuxu@ualr.edu
Department of Information Science
University of Arkansas
Little Rock 72204

**John Talburt**
jrtalburt@ualr.edu
Department of Information Science
University of Arkansas
Little Rock 72204

**Abstract:**

Large language models (LLM) are generating information at a rapid pace, requiring users to increasingly rely and trust the data. Despite remarkable advances of LLM, Information generated by LLM is not completely trustworthy, due to challenges in information quality. Specifically, integrity of Information quality decreases due to unreliable, biased, tokenization during pre-training of LLM. Moreover, due to decreased information quality issues, has led towards hallucination, fabricated information. Unreliable information can lead towards flawed decisions in businesses, which impacts economic activity. In this work, we introduce novel mathematical information quality evaluation of LLM, we furthermore analyze and highlight information quality challenges, scaling laws to systematically scale language models.

## 1 Introduction

Recently, there's been wide deployment of large language models in various fields such as medicine, law, code generator, search, with chatbots being most prominent. In domains such as software engineering, productivity is improved due to code-completion (Cruz-Benito et al. 2021) (Solaiman et al. 2019) (Moradi Dakhel et al. 2023), In writing such as grammar assistance, autocomplete, creative writing, poetry generation, question-answering systems.

Large language models use conditional probability to generate sequences of tokens for generating sentences. ChatGPT, a transformer based LLM, is creating excitement beyond measure among researchers, engineers raising questions about trust and data quality (Brants et al. n.d.). Transformers (Vaswani et al. 2017a) a language model introduced has led towards state of the art natural language processing tasks. Attention mechanisms, which make the core of language models are playing an essential role as part of the architecture in language models (Bahdanau, Cho, and Bengio 2014) (Hu 2020; Brauwers and Frasincar 2023). Attention Mechanisms enable LLM to recall, generate an accurate sequence of tokens due to their ability to focus on relevant parts of input sequence during generation of tokens. In this work, we describe trust, data quality of large language models, we describe issues around trust, data-quality, and research directions for increasing trust of language models.

In Summary, our contributions are as follows:

- We propose a novel mathematical formulation of information quality using consistency, relevance and accuracy metric, thus introducing a pipeline for LLM to evaluate information quality in natural language processing

- We explore importance of trust, data-quality for large language models, which affects economy, we highlight how unreliable data leads towards poor performance of LLM, data quality issues such as diversity of training, bias, tokenization are needed in larger datasets for LLM

- We postulate why quality of LLM is facing limitations, we find due to data quality issues, and we find datasets such as general, specialized datasets are enabling LLM to improve performance in specific domains. We also explore scaling laws like Chinchilla and Broken Neural Scaling laws, which enable LLM to systematically scale language models

- We find limitations of LLM such as hallucination, common sense reasoning, inconsistency, and we suggest research direction for improving LLM through investigating theory and principles of LLM, reducing dependence on human labellers and improving data quality

## 2. Why Information quality generated by LLM play a role in Economy?

Trust plays a central role in economic transactions, for the majority of professions in businesses. Key decisions are taken, based on the information quality available. Firstly from consumers, when they make a decision about a product, they desire reliable information to purchase, secondly, product reviews, data informs their choices.

Large language models are producing information based on training data, models such as GPT series are trained on trillions of tokens, amassing the entire internet, when LLM are used to accomplish tasks such as code-generation, tutoring students as education bots, users require them to rely and trust them. Unreliable information leads towards poor data quality, thus hampering both customers and professions to lose trust. Uncertainty hinders towards lack of growth, development. Therefore it is of urgent requirement to explore information quality, trust of large language models.

Increasing information quality might lead towards increased reliability of large language models, this might have indirect effects in the global economy such as stronger consumer confidence, effective partnership, adoption of language model technology all over the globe, increased transparency, thereby boosting economic activity of all businesses.

**3. Why are Language Models facing Information Quality issues?**

Reasons why language models are facing information quality issues is due to process of training, involving tokenization, quality of data which involves lack of diversity, bias, requiring larger dataset as LLM is being scaled, increasing in size. Large language models since introduction of transformers (Vaswani et al. 2017a) have been increasing in size, data, performance of tasks in majority of natural language processing tasks. Transformers, an encoder-decoder type of architecture, contained training data of 800 million words from Book-Corpus with 110 million parameters with 6 layers. BERT consists of encoder only architecture from Transformers, with 12 layers, 110 million parameters, trained bidirectionally.

Generative Pre-trained Transformers (GPT) consists of decoder-only architecture, with 12 layers, with 110 million parameters, trained unidirectionally, with further improvements leading towards GPT-2, containing 48 layers, 1.5 billion parameters, trained on 40GB of WebText, with GPT-3 increasing size, with 96 layers, with a total of 175 billion parameters, containing 12 attention heads, 768 dimensional states, with increasing more size, GPT-4 [(OpenAI 2023)](#) consisting of 1.8 trillion parameters with 120 layers, trained on 13 trillion tokens.

Tokenization (Toraman et al. 2023) is an important pre-processing technique for large language models, where inputs are broken down into subunits before sending to encoder-decoder layers of language models. Tokenization methods applied in language models such as character-level, byte-pair encoding (BPE), Wordpiece, morphological-level, word-level are applied in language models. Variability (Kaddour et al. 2023), token's glitching leading towards unexpected behavior, higher computational costs as it adds additional layer of training, information loss and quality.

| Method | Tokenized Text |
|---|---|
| Character-level | "L", "a", "n", "g", "u", "a", "g", "e", " ", "M", "o", "d", "e", "l", "s", " ", "a", "r", "e", " ", "i", "m", "p", "r", "e", "s", "s", "i", "v", "e" |
| BPE | "[CLS]", "Language", "Models", "are", "impress", "##ive", "[SEP]" |
| WordPiece | "[CLS]", "Language", "Models", "are", "impress", "##ive", "[SEP]" |
| Morphological | "[CLS]", "Language", "Model", "s", "are", "impress", "##ive", "[SEP]" |
| Word-level | "[CLS]", "Language", "[UNK]", "are", "[UNK]", "[SEP]" |

**Table 1: Methods of Tokenization on text, "Language models are Impressive"**

Character level tokenization takes place at character level, so there is no vocabulary for training, can represent characters, reducing model size, however it loses word level meaning and outputs long sequences, loses compositionality of words. BPE is able to handle vocabulary of subwords frequently used, and able to handle unseen words, a major disadvantage is that it is less interpretable towards morphological tokenization. Word-piece tokenizable is based on language modelling, also suffers from morphological tokenization and contains inconsistencies.

Meaningful outputs are found from morphological tokenization, which captures semantics. World level tokenization loses subword information, while capturing simple whitespace tokenization. When large language models are trained on inconsistent tokenization, vocabulary size, it impacts quality of data and performance of language models. Poor quality data (Wiseman, Shieber, and Rush 2017) results in low performance in the model, SBNATION data, scored validation perplexity of 33.34 and BLEU score of 1.78, due to noisy quality of SBNATION data. Training data of language models are massive, however the training data is frequently lacking data-diversity (Bender et al. 2021)

Language models (Kaplan et al. 2020) performance depends on scaling them, which is based on model parameters N, size of dataset D, amount of compute C for training. Based on three scale factors, N, D, C, performance has power-law relationship. Training a single LLM requires hundreds of thousands of computing hours, resulting in a cost of millions of dollars, consuming energy equal to several US families annually. Another challenge which has led towards language models suffering from data quality is bias, (OpenAI 2023) where there is systematic misrepresentation, attribution errors, factual distortion, due to several factors. One major factor is due to training data, when the training algorithm places more importance on certain features or data points. The training data is gathered from websites, books, social media platforms, conversational data, during training language models are exposed to billions of sentences, phrases, allowing them to learn relationship between words, grammar, meaning and context, when LLM is tasked with generalization, they acquire these biases which impacts results due to bias in data quality.

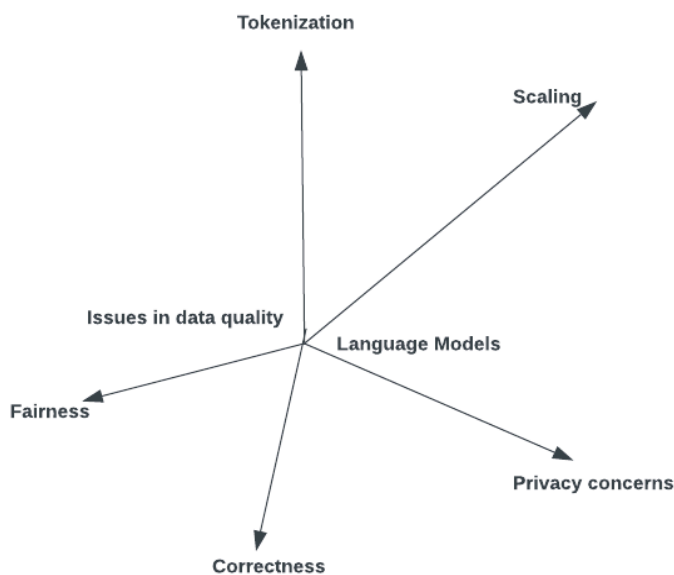

**Figure 1: Language Models Data Quality**

Four dimensions of data quality (R. Y. Wang and Strong 1996) involves intrinsic data quality, contextual data quality, representational data quality, accessibility category, where intrinsic data includes accuracy, cleanliness, addressing missing values. Contextual data quality refers to dimensions in data such as relevance, timelines, completeness, appropriateness, representational data quality refers to data which is presented in intelligible, clear to understand. Accessibility category refers to data which is obtainable, secure. For Language models, these dimensions affect the performance of large language models.

As Language models are becoming more pervasive (Batarseh and Freeman 2022; Felderer and Ramler 2021) and dependent upon AI, however quality assurance of the data and systems is necessary by data quality, which can be measured by correctness, model relevance, robustness, security, data privacy, efficiency, fairness, interpretability. These issues can be characterized by quality characters, artefact type, processes. Information Quality of language models can be measured also by consistency, relevance, accuracy metrics. We define Information quality as a function with a combination of parameters describing consistency, relevance and accuracy.

**Mathematical formulation of Information Quality Evaluation LLM:**

Motivated by issues in evaluating information quality produced from large language models, We introduce dimensions to measure information quality dimensions such as consistency, relevance and accuracy. $Let \{x_1, x_2, ..., x_n\}$ be information generated by Large Language models. To evaluate the quality of Large Language models, we propose three dimensions: accuracy, consistency and relevance. We chose three dimensions due to the following reasons. Firstly, choosing more than three might dilute the information quality dimension, secondly our formulation is domain-agnostic – together accuracy, consistency, relevance. Lastly large language models are intended to be used for specific, human expectations also known as alignment. Our formulation for evaluating Information quality aligns with the alignment of LLM.

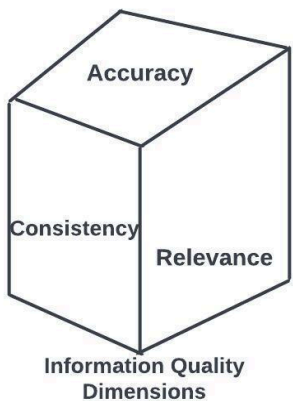

**Figure 2: Information Quality Dimensions**

We propose linear formulation of information quality due to simplicity, tunability, and normalization. Weighted sums are equal to 1.

$$IQ(L) = f((w_1 * consistency * w_2 relevance * w_3 accuracy))/\sum_{w_1}^{w_3})$$

$$\sum_{w_1}^{w_3} Iq(L) = 1$$

Here $IQ(L)$, represents information quality of the information produced by large language models, $w_1, w_2 w_3$ are context-specific, weights of the dimensions.

Accuracy, consistency, relevance are weighted combinations, which allows context-specific evaluation, which normalizes the score.

**4. State of the Art Large Language Models:**

Language has been studied by linguists, computer scientists, and statisticians. Language is human expression communicated as symbols governed by a set of grammatical rules. Researchers are investigating capable Artificial Intelligence (AI) for comprehending, grasping language. We explore state of the art language models introduced.

One significant major approach for language understanding and generation has been, Language modeling (Jurafsky and Manning 2012) (Shannon 1948), where next word probability is based on analyzing text data, various approaches are formulated in Language modeling such as n-gram, unigram, bidirectional, exponential, neural language models, continuous space. Statistical language models (SLM) were introduced in 1990s, based on markov assumption, however due to size, they suffered from curse of dimensionality, Neural (Bengio, Ducharme, and Vincent 2000) Language model introduced concept of distribution representation of words using neural networks, unified solution for NLP tasks was proposed, leading towards state of the art performance.

Recently, (Zhao et al. 2023) Transformer based language models formulated as pre-trained language models are emerging with impressive capabilities, demonstrating strong capabilities in natural language processing tasks. Abilities such as in-context learning, which have been observed, resulting from scaling these language models, the abilities are not present in smaller language models.

Transformer language models are the underlying architecture for Large Language Models (LLM) such as (Devlin et al. 2018), Bidirectional Encoder Representations from Transformers (BERT), (Radford et al. 2018) Generative Pre-trained Transformers (GPT) series (Min et al. 2021) In recent years, BERT has emerged as a powerful language model, surpassing the previous state-of-the-art models in a multitude of natural language processing tasks.

Transformers map sequence of input vectors to sequence of output vectors of same length $(a_1, a_2 ..... a_n)$ to $(b_1, b_2 ..... b_n)$. Architecturally, consisting of linear layer, feedforward network, self-attention layer, which maps two vectors of same length. Transformers (Fan et al. 2020) use attention mechanisms to capture temporal relations simultaneously processing tokens in parallel. Transformer models are able to handle long coherent outputs, enabling them to generate paragraphs. Pre-trained language models have attracted significant interest among researchers and engineers. They are trained on a large dataset and fine-tuned for specific downstream tasks. Downstream tasks such as text generation, summarization, name-entity recognition, relation extraction, and sentiment analysis.

Pre-trained models became prominent due to the introduction of ELMo (Embeddings from Language Models) (Iyyer et al., n.d.) due to the leverage of unlabelled data to create word embeddings. ELMo, word embedding method for representing sequence of words as corresponding sequence of vectors (Iyyer et al., n.d.) is based on (Hochreiter and Schmidhuber 1997) bidirectional LSTM. Pre-trained models come under three categories, Autoregressive, Masked Language, Encoder-Decoder. Popular examples of Autoregressive models are GPT, GPT-2, GPT-3, GPT-3.5 where the objective is to predict which word comes next, given previous words. A few examples of masked language models are BERT, RoBERTa, XLM-R, where the objective is to predict masked words given other words in a sequence. Recently introduced state of the art language models are masked language models – Llama, Llama 2 (Touvron, Martin, et al. 2023) (Touvron, Lavril, et al. 2023) are large language models with scale of 7 billion to 70 billion parameters. Trained on trillions of tokens. LLama is trained on publicly available dataset.

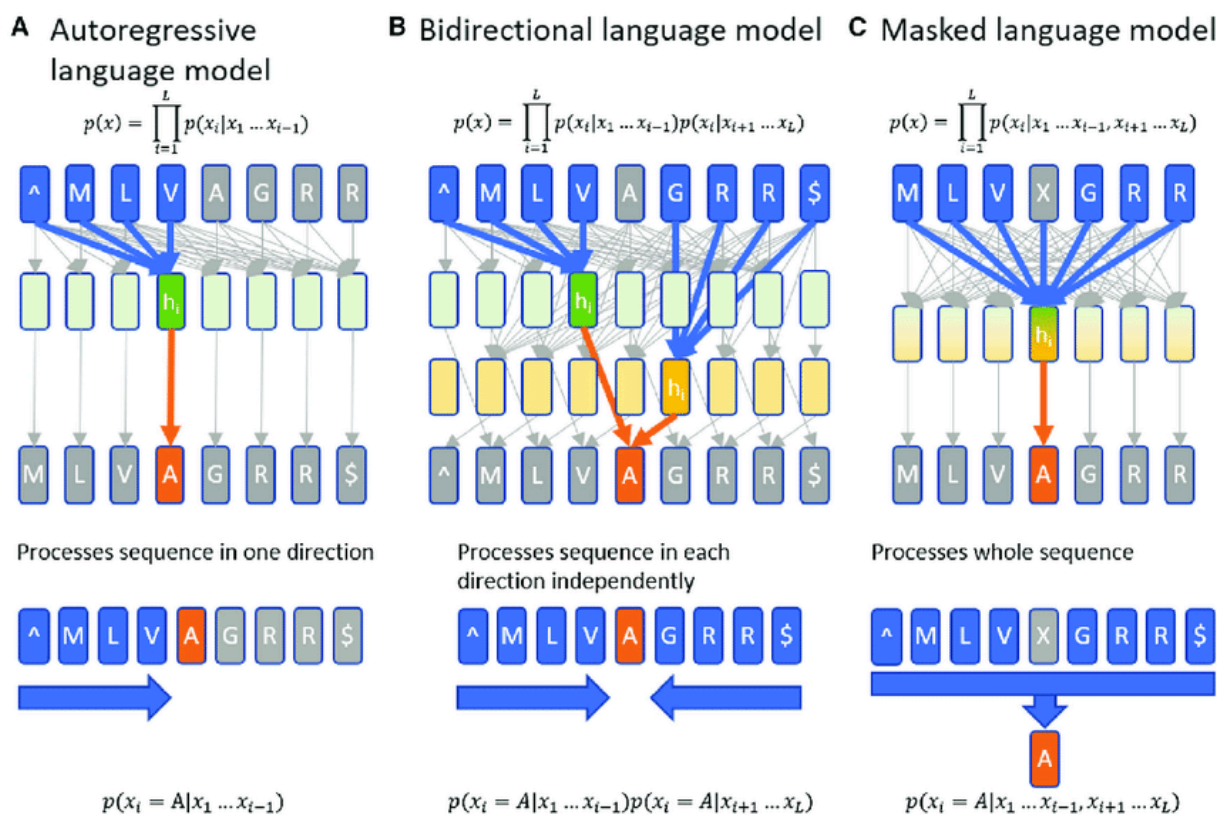

**Figure 3: Types of Pre-trained Language Model** (Bepler and Berger 2021)

BART and T5 are Encoder-Decoder models, where the objective includes corrupting a sequence and then predicting the original sequence. While researchers and engineers are interested in finding the best performing model, It is difficult to conclude which type of pre-trained model is best performing. Pre-trained models perform efficiently depending on a task. (Li et al. 2021) Autoregressive models such as GPT are highly effective for generating text, used widely for text summarization and language generation. Masked language models such as BERT excel for understanding relationships between words in a sentence and are widely used for name entity recognition and sentiment analysis. Encoder-Decoder models such as BART demonstrate efficient understanding of the structures of sentences and are widely used for tasks such as text summarization and machine translation. A transformer language model's key component is (Vaswani et al. 2017b) an attention mechanism. Attention mechanisms are trained on large datasets, from numerous sources. Due to utilization of a multi-source datasets for natural language processing tasks, attention mechanisms provide a broader, more representative sample of linguistic phenomena present in the sequence of sentences. This enables language models to learn and generalize to new examples. The datasets collected for training large language models are sourced from the entire web. These datasets are processed using techniques such as tokenization, stemming, and lowercasing for filtering out irrelevant information as higher quality data is necessary for good results. Popular benchmark datasets for running machine learning algorithms include (A. Wang et al. 2018), General Language Understanding Evaluation dataset (GLUE), which is a benchmark for performing evaluation for popular natural language processing tasks such as sentiment analysis. The Stanford Question Answering (SQuAD) (Rajpurkar et al. 2016) a commonly used question-answering dataset in natural language processing for performing evaluation in question answering, as well as Multi-Genre Natural Language Inference ((Williams, Nangia, and Bowman 2017), which is a popularly used for evaluating natural inference dataset. GPT-4 with Vision (OpenAI 2023) enables users to instruct, interact with image inputs, incorporating multimodal LLMs, which takes images, texts together as inputs (Mishkin et al. 2022), limitations remain in GPT-4V where answers might be factually incorrect, hallucinations occur in tested medical, scientific image data.

| Language Model | BERT | GPT3 | BART | ChatGPT | LLama2 |
|---|---|---|---|---|---|
| **Size** | Base: 110M Large: 340M | 175B | Base: 110M Large: 340M | 175B | 7B to 70B |
| **Training Time** | Base: 8 x V100 x 12 days<br><br>Large: 64 TPU Chips x 4 days | Months | 1 day | Months | Months |
| **Performance** | Outperforms Transformer Model | Outperforms every model | Outperforms BERT on NLG tasks | Outperforms on Conversation | Outperform LLama1 |
| **Data** | 16 GB BERT data (Book corpus + Wikipedia) 3.3 billion words | 570 GB Common crawl WebText2 Books1 Books2 Wikipedia | 160 GB | 570 GB Common crawl WebText2 Books1 Books2 Wikipedia | Larger dataset than GPT3 |
| **Method** | Bidirectional Transformer with Masked Language Modelling and Next Sentence Prediction | Decoder Only Transformer | Input text corrupted by arbitrary noise | Decoder only Transformer with RL using Human feedback | Masked Language Modeling |

**Table 2: Comparison of popular SOTA Large Language Models**

**Scaling laws of Language Models**

In the above table, Transformer based language models displayed an increase in performance with an increase in size of parameters in language models. Researchers found increasing size of dataset (Kaplan et al. 2020) improved performance in many natural language processing tasks. Moreover, not only size of training data, language model parameter size increases, training time and power consumption are also increasing, researchers have proposed architectures combining with a sparsely activated mixture of experts (Du et al. 17--23 Jul 2022) to reduce power consumption.

Due to higher power consumption, researchers investigated parameter size of language models to discover scaling laws (Hoffmann, Borgeaud, Mensch, Buchatskaya, Cai, Rutherford, de Las Casas, Hendricks, Welbl, Clark, Hennigan, Noland, Millican, van den Driessche, Damoc, Guy, Osindero, Simonyan, Elsen, Rae, et al. 2022), introduced Gopher, language modeling at scale (J. Rae, Irving, and Weidinger, n.d.). (Kaplan et al. 2020) discovered there is power law relationship between number of parameters in autoregressive language models and performance. Scaling (Kaplan et al. 2020) laws for language models can be characterized by four parameters, size of model, size of training dataset, cost of training, performance after training, N, D, C, L (number of parameters, dataset size, computing cost, loss). Chinchilla scaling, a particular scaling law states (Kaplan et al. 2020; Hoffmann, Borgeaud, Mensch, Buchatskaya, Cai, Rutherford, de Las Casas, Hendricks, Welbl, Clark, Hennigan, Noland, Millican, van den Driessche, Damoc, Guy, Osindero, Simonyan, Elsen, Vinyals, et al. 2022) for training large language models auto-regressively, one epoch with cosine learning rate, The scaling law is as follows:

$C = C_0 N D$

$L = \frac{A}{N^\alpha} + \frac{B}{D^\alpha} + L_0$

C is cost of training the model

N is number of parameters of the model

D is number of tokens in training set

L is average negative log-likelihood per token

$L_0$ represents loss of ideal generative process on test data

$\frac{A}{N^\alpha}$ captures Transformer language model with N parameters underperform ideal generative process

$\frac{B}{D^\alpha}$ captures fact model trained on D tokens underperforms ideal generative process

Chinchilla scaling law (Jared, Sam, and Tom, n.d.) given increased budget in floating point operations per second (FLOP), to achieve compute-optimal, number of model parameters (N) and number of tokens for training model (D) should be scaling in approximately equal proportion.

**Broken Scaling Laws (BSNL)**

Power laws frequently occur as distributions in many phenomena, such as frequency of words in most languages, frequency of family names, size of power outages.Power law occurs when there is a functional relationship between two quantities, where relative one change occurs in another quantity to a power of change.

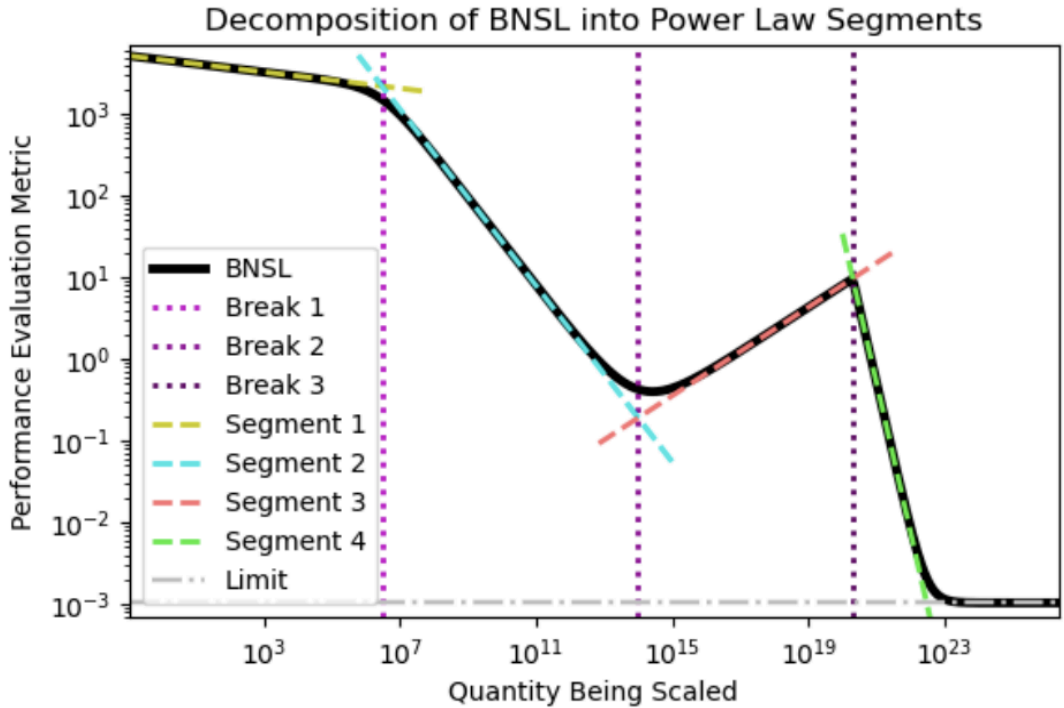

**Figure 4: Functional form of Broken Neural Scaling Law** (Caballero et al. 2022)

In language models, scaling law relates to parameters of neural networks, where neural networks are described using parameters such as size of model, size of training data, cost of training, performance of training. Smoothly broken power law which models extrapolates scaling behaviour of deep neural networks (Caballero et al. 2022), smoothly broken power law is referred as broken neural scaling law, how evaluation metric varies as amount of computation, number of model parameters, training dataset size, upstream performance. BSNL is able to extrapolate scaling behaviour, which other functional forms are incapable of expressing such as non-monotonic transitions present in scaling behaviour such as double descent, sharp inflection point.

$$y = a + (bx^{-c_0}) \prod_{i=1}^{n} \left(1 + \left(\frac{x}{d_i}\right)^{1/f_i}\right)^{-c_i * f_i}$$

Where y is performance metric, which can be used for describing accuracy

X is scale factor, which can be used to describe dataset size, compute, parameters

a, b, $c_0, c_1, c_n$ are constants

$d_1 \ldots\ldots d_n$ indicates where scaling behavior changes

$f_1 \ldots\ldots f_n$    controls sharpness of transitions

BSNL offers many advantages in neural networks. One major advantage is the ability to model expressiveness in large language models, modeling sharp, smooth breaks while allowing scale in large language models. It also allows non-monotonic relationships like double descent, accurate fit neural network scaling curves. While BSNL offers advantages, it suffers from limitations such as requiring huge data points, harder to fit, and difficult to interpret results.

## 5. Large Language Models and Data Quality for Performance:

Large language models are giving state of art performance in downstream tasks such as named-entity recognition, text generation, question-answering, translation, coreference resolution, involved in the majority of natural language processes (Brown et al. 2020; Bubeck et al. 2023).

Outstanding capabilities of LLM are due to properties such as expressivity, scalability, multimodality, memory capacity, compositionality (Bommasani et al. 2021). Due to the five properties, LLM is able to distill, accumulate knowledge from many sources, domains, organize, effective, scalable representation, flexible generalization towards novel context. GPT-3 (Brown et al. 2020), a Large language model has 175 billion parameters, which was trained on 570 gb of text data, GPT-3 demonstrated zero-shot generalization to downstream tasks, when increased in size of data and parameters from GPT-2 containing 1.5 billion parameters. Large Language Model Meta AI (Llama) (Touvron, Martin, et al. 2023) is from 7 billion to 70 billion parameters, trained on 2 trillion tokens of data on publicly available data focused on factual data.

As Large Language models are trained on large quantities of data, (Kaplan et al. 2020) where GPT based models such as WuDao (Yuan et al. 2021) is trained on 4.9 TB of multi-modal data, and scaling of dataset is only increasing due to emergent abilities of language models (Wei, Tay, et al. 2022). For training and increasing performance of LLM data quality and size is necessary. However, the available, public dataset is small compared to petabytes of consumer, business, personal data collected in industrial information, therefore there is a scalability problem of handling large datasets. Language models require both structured and unstructured data during training, (Orr et al. 2020) (Poerner, Waltinger, and Schütze 2020) demonstrated integrating structured and unstructured data can help to generalize rare concepts, improve recall and accuracy.

Moreover, effectively integrating both structured and unstructured datasets for increasing performance of large language models is a challenge. While LLM is able to be trained on unstructured data, additional steps are required for data with source containing entity or image data. In industrial scale, to integrate multi-modal dataset for text, image, more steps are required. Data privacy is another issue in integrating the data, where web crawlers are scraped for datasets without consent of users, how to preserve privacy, yet maintain high performance of large language models remain a persistent issue, where emerging regulation, guidelines ensure safe, responsible management of data.

Data Quality of Language Models: Data quality which is used to train large language models greatly affects performance (Lee et al. 2021). Data used in LLM consists of bias of many types such as social bias, false information, duplication (Blodgett and O'Connor 2017) (Fleisig et al. 2023) misinformation, which impacts training data. Due to constant change in businesses, products, data is evolving, requiring to be continuously updated and refined (Kiela et al. 2021), which has emergent abilities (Fetahu, Anand, and Anand 2015), distribution shift, meaning concept shift (Mayee Chen et al. 18--24 Jul 2021) (Oortwijn, Bloem, and Sommerauer 2021). LLM when deployed has found to undesirable behavior on critical, fine-grained sub-population of data, such as bias, which needs to be detected and mitigated, tools to detect are highly required (Hohman et al. 2018) (Goel et al. 2021)

## 6. Language Models and Datasets for Training:

Large language models are pre-trained on numerous corpus of dataset. Pre-training is a necessary step for allowing a language model to learn statistical properties, structure of human language present in data corpus. During pre-training, LLM learns the relationship between syntax, meaning, structure present between words in sentences. Vector representation of words are captured in semantic and syntactic ways through word embeddings. Word embeddings represent each word as high-dimensional vectors, which are used for optimizing for pre-training. Pre-training allows LLM to model as universal language model, which can be applied to downstream NLP tasks, that can be transferable, Pre-training in language models such as (Devlin et al. 2018) BERT involves masked language modeling, next sentence prediction, which gives BERT ability for bidirectional representation and contextual representation. Language understanding, generation are established due to pre-training (Brown et al. 2020), even demonstrating few-shot abilities. Datasets are key towards these abilities of LLM, high-quality data and how datasets are pre-processed is necessary. Datasets which are used for training can be described as general data and specialized data (Chowdhery et al. 2022; S. Zhang et al. 2022) Datasets such as multi-lingual data, scientific data, code data have enabled LLM to solve specific tasks (Taylor et al. 2022)

**General Data for LLM:** General text data contains dataset for general purpose LLM, thich includes the majority of dataset such as webpages, books, conversational text (Raffel et al. 2020). Webpage datasets contain large amount of data, which is absorbed by using (Smith et al. 2013; Patel 2020) common-crawl data, due to huge volume of internet data, LLM gains diverse capacities, however there are limitations as crawled data might contain high-quality text found in Wikipedia, and low-quality text like email spam mails. Conversational text data enables LLM to enhance the ability of LLM for conversational NLP tasks such as question-answering. Conversational datasets such as reddit corpus (Baumgartner et al. 2020) (Roller et al. 2020), social media corpus are used. Book datasets (Gao et al. 2020) provide an important source to learn longer texts in corpus, which allows LLM to model long-term dependency, generate narrative, coherent texts.

**Specialized Data for LLM:** While general datasets are used for General use LLM, specific corpus involving specialization is used to improve downstream tasks such as Multilingual text, Scientific text, Code text, which can be used for LLM based applications for specialized tasks. Specialized tasks such as code-generation, code-question-answer, scientific question-answering. PaLM, BLOOM (Chowdhery et al. 2022; BigScience Workshop et al. 2022) involves multilingual datasets which includes 46 and 122 languages with their pre-training corpus, due to multilingual dataset, PaLM, BLOOM demonstrates state of the art performance in multilingual tasks such as translation, multilingual summarization. Scientific research involves publishing findings from research, however there's growing number of scientific publications (Taylor et al. 2022), which can be incorporated into LLM, that has led towards increased performance in scientific reasoning tasks (Saier, Krause, and Färber 2023). Scientific corpus is collected from arXiv papers, scientific textbooks, physics webpages, tutorials, scholarly articles and other resources. These datasets contain mathematical formalism, symbols, and different formats, which require preprocessing so that they can be processed by language models. Code datasets have been collected, researched for program synthesis, which can be used for Pre-trained language models on Code (Piccolo et al. 2023; Mark Chen et al. 2021) (Simon 1963) GPT-J (B. Wang and Komatsuzaki 2021) and other LLM (Austin et al. 2021) has lead towards improvement in quality of the software programs. Demonstrated generated programs trained on code-specialized datasets have successfully (Mark Chen et al. 2021) passed unit test cases, solve competitive programming questions, coding datasets are collected from Stack Exchange, Github, which includes solutions and troubleshooting software errors.

**Data-Preprocessing:**

Data preprocessing for LLM involves removing noisy, redundant, irrelevant, toxic data, which affects performance. Data-juicer for LLM (D. Chen et al. 2023) involves features which allows 50 processing operations and tools, for quality filtering in low-quality data from corpus, two strategies can be applied, classifier, heuristics. Classifier identifies lower quality data and filters from the corpus, (Du et al. 17--23 Jul 2022), binary classifiers can be used with curated data such as wikipedia with positive classification.

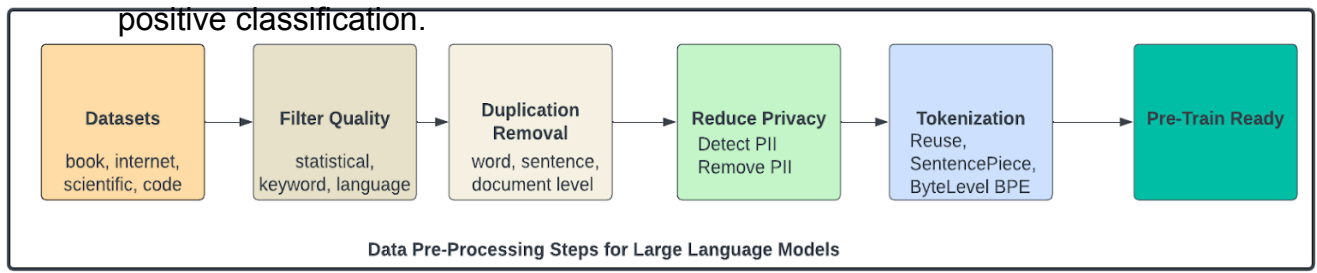

Data Pre-Processing Steps for Large Language Models

**Figure 5: Pre-training steps in Language Models**

Language models such as Gopher (J. Rae, Irving, and Weidinger, n.d.) applied a heuristic based approach to eliminate lower quality text, using well-designed rules, which filters applied on language based filtering, metric based filtering, statistic, keyword based. Language models, when applied to Tamil language tasks, can use language filters to remove non-Tamil tokens in the dataset, perplexity frequently used as evaluation for quality of language models, could be used as a metric for filtering unnatural sentences. Low quality data can be removed also using symbol to word ratio, punctuation distribution, statistical filtering, keywords such as offensive words, toxicity in sentences can be removed using keyword filtering.

**De-duplication** occurs at document-level, sentence level, dataset level resulting in lower quality data, sentences which contain repeated words, phrases must be removed as these may result in modeling repetitive patterns in language modeling. Document level duplication can be detected by using n-gram, finding similar content, dataset level duplication can be removed by finding duplicate texts from the training set. Performance of LLM greatly increases due to removed of duplicates at three levels. Protecting privacy of users is important (Carlini et al. 2022) as LLM are trained on datasets which might contain personal information. Methods such as keyword detection, such to remove personal information name, address, phone numbers can be applied to remove sensitive data contained in user-generated content from web forums.

Tokenization as described in Figure 4 is important steps, where data is preprocessed. During step of tokenization, individual tokens are segmented from raw dataset, which are used as inputs for LLM, methods such as sequence labeling with conditional random fields, word based tokenization is used, other method such as subword tokenization have been used in Transformer based models, which applies BytePair Encoding tokenization, WordPiece tokenization, Unigram tokenization. BytePair tokenization introduced as a data compression algorithm (Gage, n.d.; Sennrich, Haddow, and Birch 2015), basic symbols are iteratively combined with frequent pairs of two consecutive tokens in corpus as new tokens, as merging.

Merging is based on co-occurrence frequency of two contiguous tokens. Byte-level BPE is used to improve tokenization quality for multilingual corpus. Wordpiece tokenization is similar to Byte level BPE, instead the merge step involves selecting different criteria for merge, where in each merge, it selects the pair that leads to the most increase in likelihood of training data. Moreover other methods such as unigram tokenization are used in language models such as T5 and mBART. Pre-training data from a large mixture of data can enable LLM to acquire higher scope of knowledge, Gopher (J. W. Rae et al. 2021) uses ablation experiment on distribution of dataset to examine impact of mixed source on different tasks. High-quality data for adequately training the model results is good performance (Chung et al. 2022)

**7. Issues and Limitations of Language Models:**

Although Large language models have demonstrated state of the art performance, they face major challenges and limitations. The major issue has been the problem of hallucination (McKenna et al. 2023; Lee et al. 2018; del Campo and Leach 2022) in large language models, where given a text-corpus when LLM is prompted to produce factual text, LLM produces hallucinationed text. Hallucinations text is when generated text is in conflict with source (intrinsic) or cannot be verified by available source, (extrinsic). Due to this, LLM generates untruthful information

Hallucination has been found in all major language models, even prone in GPT-4, even occurs in multi-modal vision-based language models (Bang et al. 2023). LLM utilize knowledge in solving tasks, which lacks ability to accurately control use of internal or external knowledge. Due to hallucination, it is not recommended to deploy in real-world applications involving healthcare or other areas, which requires a high level of reliability. To mitigate this, strategies such as use of high-quality data, use of human-feedback through reinforcement learning has been applied. Applying external knowledge for checking credibility of information has reduced hallucination issue SelfCheck-GPT (Manakul, Liusie, and Gales 2023) detects hallucinations by measuring information inconsistency with sampled output, TruthfulQA, HaluEval uses LLM generated and human annotation to evaluate ability of LLM to recognize hallucination in task-specific (S. Lin, Hilton, and Evans, n.d.; Li et al. 2023) scenarios.

While LLM are trained on large quantities of dataset, another major limitation has been, when tasked with challenging recent events or knowledge, which goes beyond training dataset, LLM is limited to provide reasonable answers. Knowledge recency is an issue as cost of training is high for fine-tuning with newest information, even when fine-tuned on recent knowledge, it is likely to cause catastrophic forgetting (Kaushik et al. 2021). Recently introduced framework, EasyEdit (P. Wang et al. 2023) has been released to facilitate research of knowledge editing for LLM, how to update effectively within LLM remains an open research problem. While LLM has also demonstrated reasoning capabilities (Bubeck et al. 2023), many tasks require reasoning and relying on logical relations and evidence about factual knowledge on a particular question, chain of thought (Z. Zhang et al. 2022) prompting has been proposed for enhancing complex reasoning capacity of LLM, where intermediate reasoning steps can be manually created, automatically generated into prompts to guide LLMs to perform multi-step reasoning, reformulating tasks such as code generation improves performance. However, for tasks such as common sense reasoning, LLM and human performance is compared on Commonsense QA. LLM has accuracy of 55.9% (Wei, Wang, et al. 2022) (Chowdhery et al. 2022) (Dhingra et al. 2023), while human accuracy on dataset is about 89%, BERT performed below 10% on non-entailment category. LLM might miss or generate inaccurate intermediate steps, leading towards wrong final results. Thus, Reasoning inconsistencies are observed, even though LLM might produce correct answers, it might produce a wrong answer after a correct answer, (Yao et al. 2023) Tree of thoughts introduced mitigates this issue by empowering decision making process by exploring, self-evaluating reasoning paths. LLMs are found to have limitations when tasked with numerical computation, especially when tokens are not encountered in the dataset.

LLMs are desired to produce output, which conform to human needs and values i.e human alignment (Zhao et al. 2023; Liu et al. 2023; OpenAI 2023), where LLM can be applied and used in real-world applications, due to this, TruthfulQA (S. Lin, Hilton, and Evans, n.d.), harmlessness CrowS-Pairs (Nangia et al. 2020) datasets are used to evaluate ability of LLMs towards human-alignment, LLM consists of ability to receive feedback from external environment, perform action according to behaviour instruction, generation action plans, manipulate agents (W. Huang et al. 17--23 Jul 2022), LLM also consists of ability to manipulate tools received by API calls, such as being applied to search engine, calculator, compiler, to enhance performance (Parisi, Zhao, and Fiedel 2022). In addition to manipulating existing tools, LLM possess capability to make their own tools for specific tasks autonomously, where models independently explore, manipulate, and self-create tools, which is giving them the ability to solve real-world tasks.

LLM in addition to the above three abilities such as human alignment, tool manipulation, interaction with external environment (Gilardi, Alizadeh, and Kubli 2023), LLM is able to do data annotation, self-improvement (Gamil 2023; J. Huang et al. 2022), Evaluation methods using benchmarks are automated, however they suffer from training, test data contamination, human evaluation measure real world performance, however expensive and not reproducible, model based evaluation are efficient, however suffer from bias. (Shaikh et al. 2021) Data imbalance, a frequently occurring problem, where in the training dataset, one category of data exceeds the other category, thus causing imbalance in the dataset. Lack of enough data samples across class labels results in poor classification performance.

To align LLM with human values such as harmlessness, honesty, helpfulness, honesty, Reinforcement learning from human feedback (RLHF) is used, which uses a reward model from human feedback, implemented as a reward function to optimise agent's policy using reinforcement learning (RL) using proximal policy optimization. Human feedback is gained by asking humans to rank instance's of LLM's behaviour from outputs generated. Human feedback is executed through ranking based (Ziegler et al. 2019), question-based approach (Nakano et al. 2021) and rule based approach (Glaese et al. 2022), which Sparrow (Glaese et al. 2022) selects based on rules to test whether model-generated responses meet alignment criteria. While RLHF has mitigated hallucinations to a certain degree, it suffers from limitations such as requiring training multiple LMs as model being aligned, reward model, reference model, which requires tedious process in algorithmic and consumes so much memory. Another major limitation is that RLHF is extremely complex and time consuming to deploy.

## 8. Transformers: Architectural Foundation for Large Language Models

Transformers are the foundation for large language models (Vaswani et al. 2017a; Zhao et al. 2023). Transformers introduced as a novel architecture for neural machine translation established as state of art performance on English-German translation task. Following initial success in natural processing tasks – transformers architectures were rapidly adopted across diverse range of domains beyond natural language processing, even establishing as state of art performance in field of computer vision, in tasks such as image segmentation, multi-modal text and image generation and image recognition, using Vision-Transformers (Dosovitskiy et al. 2020). Transformers replaced convolutional, recurrent layer with attention mechanisms. Attention mechanisms were introduced to solve (Bahdanau, Cho, and Bengio 2014) fixed-length vector problem in neural machine translation (NMT). Fixed-length vector which decoder generates translation was a bottleneck. By allowing a encoder-decoder model to automatically soft-search for source sentences, which are relevant to target word, state of the art performance on NMT was achieved.

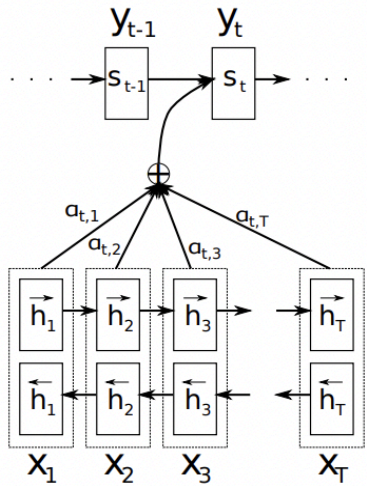

Figure 1: The graphical illustration of the proposed model trying to generate the $t$-th target word $y_t$ given a source sentence $(x_1, x_2, \ldots, x_T)$.

**Figure 6: Attention Mechanism** (Bahdanau, Cho, and Bengio 2014)

Attention mechanisms allows Transformers to focus on relevant parts of the input, when generating outputs, this enables them to capture dependencies, and allows for parallelization. However, attention mechanisms face limitations such as requiring higher computation performance, lacking deeper reasoning, semantic understanding, and requiring a large dataset. Transformers consist of *encoder-decoder architecture* (Vaswani et al. 2017b), where encoder maps input sequence of symbol representations x to sequence of continuous representation z. Given a step z, decoder generates output sequence y of symbols one element at a time. For each step, the model is auto-regressive, depending on the previous element at a time, for generating next. Architecturally transformers contain self-attention, point-wise, fully connected layers for encoder and decoder.

Each layer consists of two sub-layers, a multi-head self-attention mechanism, second is position wise fully connected, feed-forward network, with a residual connection around each two sub-layers, layer normalisation, the output of the sub-layer is $LayerNorm(x + Sublayer(x))$, $Sublayer(x)$ is a function of the sublayer itself. Sublayers give product outputs of dimensions, $d_{model} = 512$.

Decoder consists of 7 identical stack layers, in addition to two sub-layers in the encoder, decoder inserts a third sub-layer, which performs multi-head attention over the output of the encoder stack. The output of the Transformer model is a sequence of vectors with index to input tokens.

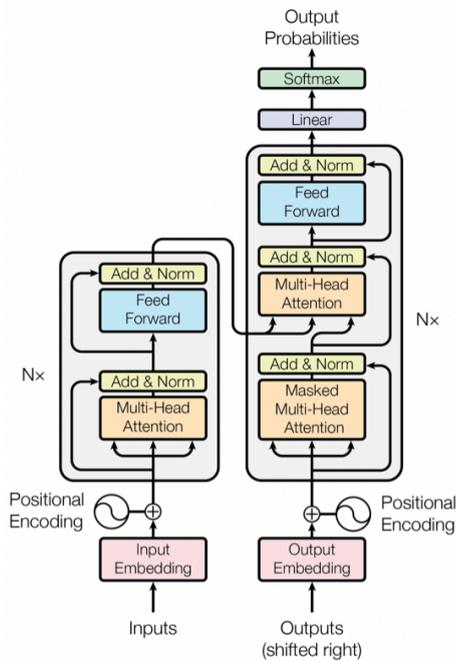

**Figure 7: Transformer model** (Vaswani et al. 2017b)

Transformers (T. Lin et al. 2022) faces issues such as requiring high computational complexity due to relying on Self-attention. Furthermore, Transformers are unable to understand long-range dependencies in language due to the chunking of texts. The text has to be split into segments, before encoders can gain them as inputs, leading toward loss of understanding of a sentence, because semantic boundaries are not taken into consideration for the entire sentence.

**What is required to make Large Language Models better?**

We witnessed progress of large language models achieving state of the art performance in natural language processing, however the challenges remain, which could further performance, understanding of large language models. Performance of language models significantly is affected by dataset quality from pre-training, adaptation tuning where data collected is fine-tuned for specific domain, utilization is where LLM is applied to various tasks such as question-answering, summarization, translation, and evaluating performance of LLM.

Theory and Principles of underlying mechanisms of LLM is largely mysterious to researchers (Zhao et al. 2023), where how information is distributed, organized, utilized in language models is not understood properly. Scaling has allowed LLM to endow with emergent abilities in unexpected ways, however emergent abilities are mysterious as well, as when and how emergent abilities appear is not understood. Architecturally, stacked multi-head self attention layers has been how LLM is deployed, sparse attention has been used in (Brown et al. 2020), which has demonstrated improving performance and efficiency, by exploiting intrinsic redundancy, capturing dynamically changing attention weights. However, catastrophic forgetting (Korbak et al. 17--23 Jul 2022) is observed, which remains a challenge in neural networks, as originally learned knowledge during training in LLM becomes damaged, affecting performance and abilities of LLM. Due to LLM (BigScience Workshop et al. 2022) (Zeng et al. 2022) being sensitive to data quality, systematic approaches for optimizing, factors of model effectiveness, efficiency optimization, training stability is required for economical reasons. Reinforcement Learning with Human Feedback (RLHF) (Ziegler et al. 2019; OpenAI 2023) has reduced hallucinations, toxicity generation from language models, However major limitation of RLHF is reliance on high-quality human feedback requiring professional labelers, which is difficult to implement in practice. Therefore, reducing human labelors with guaranteed data quality is required and needed.

## 9. Conclusion

With the rise of Large Language models deployed as ChatGPT, Claude, Bard, it is imperative to investigate processes towards building trustable language models. We have reviewed why information quality of data plays a key role in economy, highlighting information quality issues, investigating why language models performance is decreasing due to process of training, involving tokenization, quality of data which involves lack of diversity, bias, requiring larger dataset as LLM is being scaled, increasing in size. Moreover, we also explored state of the art language models with masked language models, autoregressive, and bidirectional models exploring performance. We also explored scaling laws of large language models with chinchilla and broken scaling laws, which helps researchers and engineers to scale systematically in a scientific way.

In addition to scaling laws, we explored information quality issues which LLM's performance is limiting, Information quality issues such as bias of many types such as social bias, false information, duplication, misinformation is a challenge in large language models, to improve performance of large language models, generalized dataset, and specialized dataset are being used, We found data pre-processing steps such as filter quality, duplication removal, reduce privacy, tokenization, greatly affect performance of large language models. We also found that to make LLM better in performance, theory and principles of LLM is to be investigated, reducing human labellers, and requiring high-quality data is recommended. A promising research direction is to increase high-data quality and investigate theory and principles of LLM.